%% file: main.tex
\documentclass[conference]{IEEEtran}
\usepackage{cite}
\usepackage{amsmath,amssymb,amsfonts}
\usepackage{algorithmic}
\usepackage{graphicx}
\usepackage{textcomp}
\usepackage{xcolor}
\usepackage{fancyhdr}
\usepackage[hyphens]{url}

\def\BibTeX{{\rm B\kern-.05em{\sc i\kern-.025em b}\kern-.08em
    T\kern-.1667em\lower.7ex\hbox{E}\kern-.125emX}}

\fancypagestyle{firstpage}{
  \fancyhf{}

  \fancyhead[C]{\normalsize{%ISCA 2020 Submission
      }} 
  \fancyfoot[C]{\thepage}
}

\title{High Throughput Matrix-Matrix Multiplication between Asymmetric Bit-Width Operands}
\author{}
\author{Dibakar Gope, Jesse Beu, and Matthew Mattina\\
Arm ML Research Lab\\
{\tt\small \{dibakar.gope, jesse.beu, matthew.mattina\}@arm.com}
}
\begin{document}
\maketitle
\thispagestyle{firstpage}
\pagestyle{plain}

%%%%%% -- PAPER CONTENT STARTS-- %%%%%%%%

\begin{abstract}

\input{abstract}

Index Terms - Convolutional Neural Networks, Inference, Matrix Multiplication, Hardware Accelerators, GEMM, Systolic Array
\end{abstract}

\input{intro}
\input{related_work}
\input{int8int8_matmul}

\input{int8int4_matmul}
\input{evaluation}
\input{ml_accelerator}
\input{conclusion}

%%%%%%% -- PAPER CONTENT ENDS -- %%%%%%%%

%%%%%%%%% -- BIB STYLE AND FILE -- %%%%%%%%
\bibliographystyle{IEEEtranS}
\bibliography{refs}
%%%%%%%%%%%%%%%%%%%%%%%%%%%%%%%%%%%%

\end{document}

%% file: abstract.tex
Matrix multiplications between asymmetric bit-width operands, especially between $8$- and $4$-bit operands are likely to become a fundamental kernel of many important workloads including neural networks and machine learning. While existing SIMD matrix multiplication instructions for symmetric bit-width operands can 
%handle 
support operands of mixed precision by zero- or sign-extending the narrow operand to match the size of the other operands, they cannot exploit the benefit of narrow bit-width of one of the operands. We propose a new SIMD matrix multiplication instruction that uses mixed precision on its inputs ($8$- and $4$-bit operands) and accumulates product values into narrower $16$-bit output accumulators, in turn allowing the SIMD operation at $128$-bit vector width to process a greater number of data elements per instruction to improve processing throughput and memory bandwidth utilization without increasing the register read- and write-port bandwidth in CPUs. The proposed asymmetric-operand-size SIMD instruction offers $2\times$ improvement in throughput of matrix multiplication in comparison to throughput obtained using existing symmetric-operand-size instructions while causing negligible ($0.05\%$) overflow from $16$-bit accumulators for representative machine learning workloads. The asymmetric-operand-size instruction not only can improve matrix multiplication throughput in CPUs, but also can be effective to support multiply-and-accumulate (MAC) operation between $8$- and $4$-bit operands in state-of-the-art DNN hardware accelerators (e.g., systolic array microarchitecture in Google TPU, etc.) and offer similar improvement in matrix multiply performance seamlessly without violating the various implementation constraints. We demonstrate how a systolic array architecture designed for symmetric-operand-size instructions could be modified to support an asymmetric-operand-sized instruction.

%% file: intro.tex
\section{Introduction}
\label{intro}

Use of deeper and wider convolutional neural networks (CNNs) has led to impressive predictive performance in many machine learning applications, such as image classification, object detection, semantic segmentation, etc. However, the large model size and associated computational inefficiency of these deep neural networks often make it impossible to run many realtime machine learning tasks on resource-constrained mobile and embedded devices, such as smartphones, AR/VR devices, etc. One particularly effective approach has been the use of model quantization to enable this size and computation compression of CNN models. Quantization of model parameters to sub-byte values (i.e. numerical precision of $\leq$ $8$ bits), especially to $4$-bits has shown minimal loss in predictive performance across a range of representative networks and datasets in recent works. 
As a result, some heavily quantized machine learning models may use kernel weights which have fewer bits than the corresponding activations which they are to be multiplied with.  For example, there is an increasing interest in using $4$-bit weights and $8$-bit activations, which means that matrix multiplications between $4$-bit weights and $8$-bit activations are likely to become a fundamental kernel of many important workloads including neural networks and machine learning, although such multiplications may also be useful for other purposes.
This is evident by the increasing interest and successful development of a large number of novel machine learning and linear algebra techniques~\cite{Banner_NeurIPS2019, Louizos2019,Yang2019,Gong_2019_ICCV,Jung_2019_CVPR} to preserve the predictive performance of deep neural networks with $4$-bit weights and $8$-bit activations in recent years.
However, in $4$-bit-weight networks, the weights are encoded by $4$ bits, while the activation matrices are represented by more bits (e.g., $8$ bits in this example, although other examples could have larger activations).  This creates a read width imbalance between $4$-bit weights, $8$-bit activations and outputs (accumulators) compared to previous technology.  Ideally, we would like to sustain matched vector width of read and write operands while exploiting $4$-bit weights for the best performance. In other words, we would like to utilize the full bandwidth of read and write ports while exploiting $4$-bit weights for the best performance.  

%Using $4$-bit instead of $8$-bit encoding for the weight matrix argument of a matrix multiplication instruction (or other similar ops) means twice as many values can be accessed for the same number of bits; that’s by design and an intended consequence in order to get a speedup. Subsequently, part of the matrix multiplication hardware can be reused to do $2\times$ more multiplies of narrower width, and change the matrix architecture of the narrower argument to be twice as wide to use all the bits available.

While existing instructions with a same operand size in both first and second operands already would support application to operations involving narrower data values for the second operand, they will not be able to exploit the narrower bit-width of the second operand for improving the MAC throughput of matrix multiply operation.

In contrast, by implementing an asymmetric-operand-size matrix multiplication instruction (or other similar operations) 
%using $4$-bit elements instead of $8$-bit elements for the operand used for the weight matrix, 
using $4$-bit instead of $8$-bit encoding for the weight matrix, twice as many values can be accessed for the same number of bits
%from memory per load instruction 
– this is by design and an intended consequence in order to get a speedup.  Subsequently, part of the matrix multiplication hardware can be reused to do twice as many multiplies of narrower width, and the matrix architecture based on the narrower argument can be twice as wide to use all the bits available. 
While this improves the MAC throughput of the SIMD asymmetric-operand-size matrix multiply operation at $128$-bit vector width by $2\times$, 
%While this can offer 2x improvement in MAC throughput involving 4-bit weights and 8-bit activations in future CPUs,
the accumulation of product between $8$- and $4$-bit operands into $32$-bit accumulators doubles the register read- and write-port bandwidth requirement of the output matrix owing to larger accumulator matrix. This is overcome by accumulating products into $16$-bit accumulators. While this reduces the number of spare digits for carries, in practice for many common workloads (e.g., representative deep neural networks) overflows still do not occur often and so the concerns about overflows from $16$-bit accumulators occurring too often are misplaced.

%Our proposal addresses the following challenges created by multiplying 4-bit weights with 8-bit activations.

To summarize, we make the following contributions.

\begin{itemize}

\item To the best of our knowledge, this is the first work to propose a SIMD matrix multiply operation between asymmetric bit-width operands that offers $2\times$ increase in MAC throughput 
%without increasing the register read- and write-port bandwidth requirements 
without violating the register vector width requirements in CPUs while observing negligible ($0.05\%$) overflow from $16$-bit accumulators for ResNet18-like architectures on ImageNet dataset.

\item This SIMD instruction addresses the challenges created by mismatch between the read bandwidth (vector width) of $4$-bit weights, $8$-bit activations, and write bandwidth of $16$-bit accumulators.

\item The asymmetric-operand-size matrix multiply operation can be seamlessly integrated into a DNN accelerator (e.g., systolic array in Google TPUs, etc.) designed for symmetric-operand-size operation to achieve $2\times$ improvement in MAC throughput without violating the associated implementation constraints (e.g., the size of operand buffers and accumulator buffers).

\end{itemize}

%% file: related_work.tex
\section{Background and Related Work}
\label{sec:related_work}

In recent years, numerous research efforts have been devoted to quantizing neural network architectures 
to sub-byte values while preserving the accuracy of full-precision model~\cite{strassennets2018,Louizos2019,Yang2019,Gong_2019_ICCV,Jung_2019_CVPR,Zhuang_CVPR2019,Zhu_CVPR2019,Sun2019,LQ-Nets2018,Guo2017,GopeCVPR2020,GopeMLSys2019,TernaryMobileNetstinyMLSummit2020}. 
Furthermore, several approaches were proposed on developing compressed neural networks through the use of weight pruning~\cite{DeepCompression2016}, tensor decomposition~\cite{Jaderberg2014,HMD_EMC2NeurIPS2019, HMD_EMC2ISCA2019,Kronecker_NeurIPS2019Submission}, compact network architecture design, etc.
Learning quantization for numerical precision of $4$-bits has been shown to be effective in recent works~\cite{Banner_NeurIPS2019, Louizos2019,Yang2019,Gong_2019_ICCV,Jung_2019_CVPR}, in turn creating demand for efficient execution of matrix multiplication kernel between $4$-bit weights and $8$-bit activations on existing CPUs and DNN hardware accelerators.
%facilitating their practical use in existing processors
However, the mismatch between read bandwidth of $4$-bit weights, $8$-bit activations, and write bandwidth of accumulators poses major obstacles in implementing such an instruction for matrix multiplication hardware in CPUs and DNN hardware accelerators. 
The use of existing instructions (that execute MAC operations between symmetric bit-width operands) to perform such matrix multiplication between asymmetric bit-width operands will not be able to fully exploit the benefit of $4$-bit weight quantization.
%On the other hand, failure to match the read and write bandwidth of vector operands by a new MAC/SIMD instruction.
On the other hand, failure to match the vector width of weights, activations, and accumulators by a matrix multiply instruction will either under-utilize expensive CPU resources (e.g., register file port bandwidth, etc.) or require significant increase in the 
%changes to the compute and storage resources/compute and 
DNN hardware accelerator resident SRAM resources (e.g., size of accumulator buffers, etc.) to realize any throughput benefit from $4$-bit quantization.
%inappropriate size of may cause
%inappropriate use of accumulator buffer resource
None of the recent works on $4$-bit model quantization reports performance benefit on either existing CPUs or hardware accelerators.

%% file: int8int8_matmul.tex
\section{Overview of Matrix Multiplication between Symmetric Bit-Width Operands}
\label{sec:int8int4_matmul}

Traditionally, the kernel weights would have the same number of bits as the corresponding activations which they are to be multiplied with.  For example, it may be common for each activation value and kernel weight to comprise $32$ bits, $16$ bits or $8$ bits, with identical sizes for the activation and kernel values.

\begin{figure}[!tbp]
  \centering
  \includegraphics[width=1.0\linewidth]{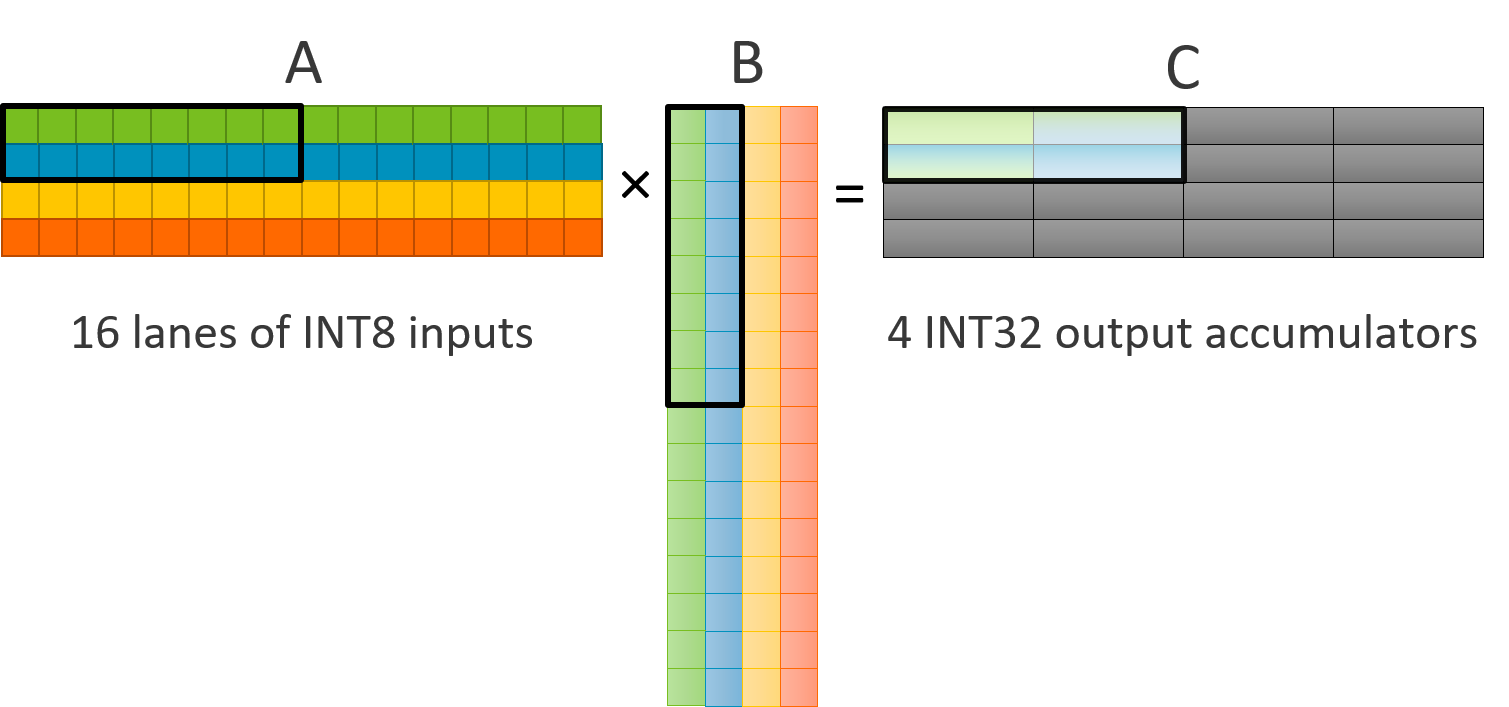}
  \caption{
  %$4\times4$ output software tiling with $2\times2$ output HW/instruction tiling through the matrix multiply instruction. 
  The $2\times2$ output cells are partially computed through reuse via A$_{top}$$\times$B$_{top}$, A$_{top}$$\times$B$_{bottom}$, A$_{bottom}$$\times$B$_{top}$, A$_{bottom}$$\times$B$_{bottom}$, improving the overall load-to-compute ratio.  In this example, $8$ steps are required to fully compute C.}
  \label{fig:int8int8matmul}
\end{figure}

Figure~\ref{fig:int8int8matmul} shows an example of implementing this matrix processing using a symmetric-operand-size matrix multiplication instruction which acts on first and second operands with identical data element sizes.  In this example, the input activations and weights both comprise $8$ bits, so the result of any single multiplication operation on two $8$-bit values will be $16$-bits wide, and as machine learning processing requires the products of two or more different pairs of activations/weights to be added together (and possibly accumulated with previous elements calculated by earlier instructions), then to avoid loss of accuracy due to overflow, the $16$-bit results are typically accumulated into $32$-bit elements in the result matrix C. This means in a vector architecture for a same-element-size implementation the input-to-output width ratio of $4:1$ works well.  

%However, an additional source of improved performance is matrix element reuse.  
Furthermore, an additional source of performance improvement is matrix element reuse.  Typically tiling (blocking) is used in software but can also be applied at an instruction level as well as seen in the proposed matrix multiplication instruction (demonstrated in Figure~\ref{fig:int8int8matmul}) by packing 2D matrices
%--rather than 1D vectors--into vector registers.
in vector registers.
As shown in Figure~\ref{fig:int8int8matmul}, the registers can be loaded with a larger number of data elements than can be processed by a single instruction, so that the elements loaded by a single set of load operations can be reused across multiple instructions in different combinations.  The portions of the activation and weight matrices indicated using the box in Figure~\ref{fig:int8int8matmul} represent the portions processed by a single matrix multiplication instruction (e.g. each portion corresponds to a sub-matrix of $2\times8$ elements of the $4\times16$-element matrix structure loaded into the registers), and the matrix multiplication instruction generates a $2\times2$ output cell within the output matrix C (each element of the $2\times2$ cell comprising a $32$-bit element).  The output of one instance of the matrix multiplication instruction only generates a partial value for that output cell – in this case corresponding to the multiplication of A$_{top}$ and B$_{top}$ shown in Figure~\ref{fig:int8int8matmul}. The final value for the output cell is computed across multiple matrix %multiply-and-accumulate 
MAC instructions by adding the results of corresponding elements derived from matrix multiplications of A$_{top}$ $\times$ B$_{top}$, A$_{top}$ $\times$ B$_{bottom}$, A$_{bottom}$ $\times$ B$_{top}$ and A$_{bottom}$ $\times$ B$_{bottom}$. The other output cells within the output matrix C can then be performed through similar calculations using different pairs of rows and columns from the loaded activation and weight matrix structures.  By reusing the same set of inputs for multiple instructions, this can improve the overall load-to-compute ratio compared to an approach where separate load operations are required to load the operands for each individual instruction. 

%Assuming $8$-bit weights and activations, the maximum possible result of any single multiplication operation is $16$-bits wide.  Due to the accumulative nature of a matrix multiplication operation, these $16$-bit results are typically accumulated in a $32$-bit register.  This means in a vector architecture the input to output width ratio of $4$-to-$1$ works well.
%(see the Dot instruction). 

%% file: int8int4_matmul.tex
\section{Matrix Multiplication between Asymmetric Bit-Width Operands}
\label{sec:int8int4_matmul}

%Assuming $4$-bit weights and $8$-bit activations, the maximum possible result of any single multiplication operation is $12$-bits wide.  Due to the accumulative nature of a matrix multiplication operation, these $12$-bit results can be accumulated into a $16$-bit accumulator register.  Furthermore, $4$-bit weights can improve the virtual bandwidth // Virtual BW? Aren't we talking about capacity of the RF here? (vector width) of register file by storing larger weight sub-matrices in the same limited-size register file.  An instance of this concept has been captured by the instruction proposal, with $128$-bit vector width shown in Figure~\ref{fig:int8int4matmul_vectorwidth128}, where register B that once held a $8\times2$ sub-matrix can now hold a $8\times4$ sub-matrix using a specialized register format.

If a quantized neural network with $4$-bit weights is executed using symmetric-operand-size matrix multiplication instructions similar to those shown in Figure~\ref{fig:int8int8matmul}, then the $4$-bit weights stored in memory could be loaded into a number of $8$-bit elements within the B operand registers, with each $4$-bit weight value from memory sign-extended or zero-extended to fill the remaining $4$ bits of each $8$-bit element of the B operand registers.  This would mean that the $4$-bit weights would not be packed contiguously into the input registers but would be dispersed into a number of non-contiguous $4$-bit chunks with gaps between them corresponding to the locations of the sign-extension or zero-extension.  Having extended the $4$-bit weights from memory into $8$-bit elements, the matrix multiplication could be performed in the same way as described above for Figure~\ref{fig:int8int8matmul} to generate four $32$-bit output accumulator values per instruction (based on the multiplication of $16$ ($2\times8$) lanes of $8$-bit activations and $16$ ($8\times2$) lanes of $8$-bit weights (expanded from the $4$-bit weights in memory)).  Hence, while this approach would allow the storage overhead of storing the weights in memory to be reduced compared to an approach using $8$-bit weights, the processing throughput 
%and memory bandwidth costs 
cost would be the same, as the number of elements processed 
%per load instruction or 
per matrix multiply instruction would still be the same as in Figure~\ref{fig:int8int8matmul}.

In contrast, by implementing an asymmetric-operand-size matrix multiplication instruction using $4$-bit elements instead of $8$-bit elements for the operand used for the weight matrix, twice as many values can be accessed from memory per load instruction – this is by design and an intended consequence in order to get a speedup.  Subsequently, part of the matrix multiplication hardware can be reused to do twice as many multiplies of narrower width.
%and the matrix architecture based on the narrower argument can be twice as wide to use all the bits available.

\begin{figure}[!tbp]
  \centering
  \includegraphics[width=0.8\linewidth]{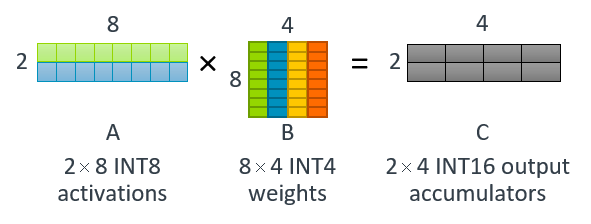}
  \caption{Matrix-matrix multiplication between $8$- and $4$-bit values at $128$-bit vector width while accumulating into $16$-bit output accumulators.}
  \label{fig:int8int4matmul_vectorwidth128}
\end{figure}

%Hence, Figure~\ref{fig:int8int4matmul_vectorwidth128} shows, for comparison, processing of $8$-bit activations and $4$-bit weights in an approach supporting a asymmetric-operand-size instruction similar to shown in Figure 2, where the second operand has data elements contiguously packed into registers with a smaller data element size than the data element size of the elements of the first operand. 
Figure~\ref{fig:int8int4matmul_vectorwidth128} shows the proposed asymmetric-operand-size matrix-matrix multiplication instruction processing $8$- and $4$-bit operands. The second operand has data elements contiguously packed into registers with a smaller data element size than the data element size of the elements of the first operand.
%Assuming $4$-bit weights and $8$-bit activations, the maximum possible result of any single multiplication operation is $12$-bits wide. 
The maximum possible result of any single multiplication operation between $8$- and $4$-bit operands is $12$-bits wide. Due to the accumulative nature of a matrix multiplication operation, these $12$-bit results can be accumulated into a $16$-bit accumulator register.  Furthermore, $4$-bit weights can improve the virtual bandwidth (vector width) of register file by storing larger weight sub-matrices in the same limited-size register file.  For example, with $128$-bit vector width shown in Figure~\ref{fig:int8int4matmul_vectorwidth128}, the B input operand register corresponding to B$_{top}$ that once held a $8\times2$ sub-matrix of $8$-bit elements can now hold a $8\times4$ sub-matrix of $4$-bit elements.  
Hence, in the example of Figure~\ref{fig:int8int4matmul_vectorwidth128} the first operand A comprises the same $2\times8$ sub-matrix of $8$-bit activations as is represented by the portion A$_{top}$ in Figure~\ref{fig:int8int8matmul}, but the second operand B comprises a sub-matrix of $8\times4$ $4$-bit weights and so corresponds to the top half of the matrix structure B shown in Figure~\ref{fig:int8int8matmul} (rather than only comprising B$_{top}$).  Hence the number of input elements in the second operand B that can be processed in one instruction is twice as many as in the symmetric-operand-size instruction shown in Figure~\ref{fig:int8int8matmul}.  Similarly, the portion of the result matrix generated by an asymmetric-operand-size matrix multiply operation shown in Figure~\ref{fig:int8int4matmul_vectorwidth128} includes twice as many elements as the portion generated by a symmetric-operand-size operation shown in Figure~\ref{fig:int8int8matmul}.  The instruction in Figure~\ref{fig:int8int4matmul_vectorwidth128} generates a $2\times4$ matrix of $16$-bit result elements, instead of generating a $2\times2$ matrix of $32$-bit elements, but can still use registers of the same size as Figure~\ref{fig:int8int8matmul}.  
Hence, while the symmetric-operand-size matrix multiply instruction shown in Figure~\ref{fig:int8int8matmul} multiplies $8$-bit activations by $8$-bit weights to generate $32$-bit output accumulators, the asymmetric-operand-size instruction shown in Figure~\ref{fig:int8int4matmul_vectorwidth128} multiplies $8$-bit activations by $4$-bit weights to generate $16$-bit output accumulators instead.  This means that the asymmetric-operand-size instruction is able to process twice as many inputs and generate twice as many outputs per instruction as opposed to the symmetric-operand-size instruction. 
%a conventional processor supporting a symmetric-operand-size instruction.

Another advantage is that as it is not necessary to zero-extend or sign-extend the narrower weights stored in memory when loading them into registers, which makes load processing simpler, and also means that the full read or write port bandwidth supported to match the register size used is available for loading the $4$-bit weights (rather than needing to artificially limit the read or write bandwidth used for an individual load instruction to half that represented by the register size to allow for the zero-/sign-extension).  Hence, support for this instruction can speed up the processing of quantized machine learning networks that use mixed precision on its inputs.

One potential challenge for widespread acceptance of an instruction like this would be overflow violations in the relatively narrow accumulators. While the matrix multiply operation in Figure~\ref{fig:int8int8matmul} uses $32$-bit accumulators to accumulate $16$-bit products resulting from multiplication of two $8$-bit operands, and so has $16$ bits spare to accommodate carries before any risk of overflow occurs, the asymmetric-operand-size operation in Figure~\ref{fig:int8int4matmul_vectorwidth128} uses $16$-bit accumulators instead to accumulate $12$-bit products,
%resulting from multiplication of an $8$-bit element and a $4$-bit element,
so there are only $4$ bits spare for accommodating carries before there is a risk of overflow.
In the worst case, only $32$ $12$-bit products resulting from signed multiplication of $8$- and $4$-bit values %($+127$ $\times$ $-8$ $=$ $-1016$) 
($+127 \times -8 = -1016$)
can be accumulated into a $16$-bit ($-32768$ to $32767$) register before overflowing.  While this would be fine for a single instance of the instruction, typical use cases reuse a stationary accumulator register over multiple instances of the instruction within a loop.

%% file: evaluation.tex
\section{Experiments and Results}
\label{sec:evaluation}

\begin{figure}[!tbp]
  \centering
  \includegraphics[width=1.0\linewidth]{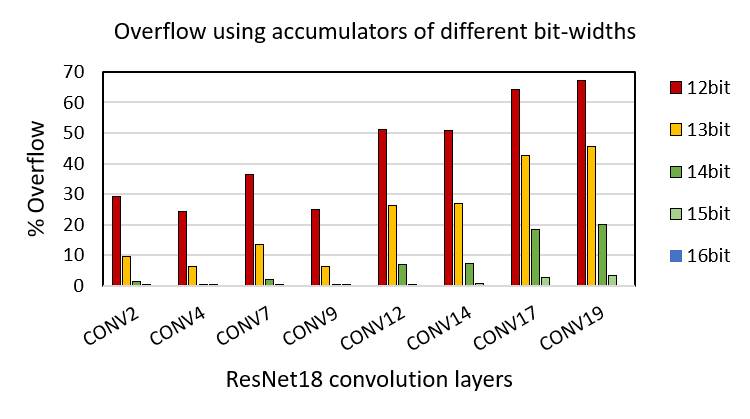}
  \caption{Overflow (\% of accumulation operation causing overflow) observed while using accumulators of different bit-widths for performing high throughput matrix multiplication between $8$-bit activations and $4$-bit weights of ResNet18 architecture.}
  \label{fig:resnet18_accum_overflow}
\end{figure}

In order to observe the amount of overflow that happens in practice while using $16$-bit accumulators for performing matrix multiplication between $8$-bit activations and $4$-bit weights in our proposal, test data from the ImageNet dataset was fed to the ResNet18 architecture where activations and weights are quantized to $8$-bit and $4$-bit respectively. For $16$-bit width of accumulator, almost non-existent ($0.05\%$) overflow (\% of accumulation operation causing overflow while generating the output activations of each layer) is observed as shown in Figure~\ref{fig:resnet18_accum_overflow} and Table~\ref{table:resnet18_accum_overflow_at_16bit}.  Figure~\ref{fig:resnet18_accum_overflow} shows the percentage of accumulation operations causing overflow observed while using accumulators of different bit-widths for performing high throughput matrix multiplication between 8-bit activations and 4-bit weights of the ResNet18 architecture.  Table~\ref{table:resnet18_accum_overflow_at_16bit} shows the overflow observed while using a $16$-bit accumulator for performing matrix multiplication between $8$-bit activations and $4$-bit weights.
Table~\ref{table:resnet18_ops_count} shows the number of matrix MAC operations ($C_{in} \times w \times h$) performed for generating each output element of different layers of the ResNet18 architecture, where $C_{in}$ is the number of input channel values, and $w$ and $h$ are the width and height of each kernel array. 
%In order to observe the amount of overflow that happens in practice while using $16$-bit accumulators for performing matrix multiplication between $8$-bit activations and $4$-bit weights in our proposal, test data from ImageNet dataset is fed to the ResNet18 architecture where activations and weights are quantized to $8$-bit and $4$-bit respectively. For $16$-bit width of accumulator, almost non-existent ($0.05\%$) overflow (\% of accumulation operation causing overflow while generating the output activations of each layer) is observed as shown in Figure~\ref{fig:resnet18_accum_overflow} and Table~\ref{table:resnet18_accum_overflow_at_16bit}.

Table~\ref{table:resnet18_accum_overflow_at_16bit} and Table~\ref{table:resnet18_ops_count} show that in practice overflow only happens in the largest of neural network layers (which are falling out of favour compared to more efficient modern architectures) where over $2000$ multiplication results are accumulated into each $16$-bit accumulator result.  This demonstrates that in the common case overflow for $16$-bit accumulators is very rare.
%Table~\ref{table:resnet18_ops_count} shows that in practice overflow only happens in the largest of neural network layers (which are falling out of favor vs more efficient modern architectures) where over $2000$ multiplication results are accumulated into each $16$-bit accumulator result.  This demonstrates that in the common case overflow for $16$-bit accumulators is very rare. 

Hence, the matrix multiplication operation between asymmetric bit-width operands proposed in this work is not expected to cause significant difficulties concerning the occurrence of overflow.  If overflow detection is desired, making the overflow \textit{sticky} (in that the max negative or positive value does not change once it is %reached/
overflowed) can enable a simple error detection routine as well by scanning the outputs for any $-MAX\_VALUE$ and $+MAX\_VALUE$ results.  Additionally, since machine learning workloads are tolerant to such numerical errors, in most use cases the sticky max values can just be used directly in the next stage of compute without any checking routine.
%If overflow detection is desired, making the overflow ‘sticky’ (in that the max negative or positive value does not change once it is reached/overflowed) can enable a simple error detection routine as well by scanning the outputs for any -MAX\_VALUE and +MAX\_VALUE results.  Additionally, since machine learning workload are tolerant to such numerical errors, in most use cases the sticky max values can just be used directly in the next stage of compute without any checking routine. 

\begin{scriptsize}
\begin{table}[h!]
  \centering
  \caption{Overflow (\% of accumulation operation causing overflow) observed while using $16$-bit accumulator for performing high throughput matrix multiplication between $8$-bit activations and $4$-bit weights of ResNet18 architecture.}
  \label{table:resnet18_accum_overflow_at_16bit}
  \begin{tabular}{|l|l|}
    \hline
    \textbf{ResNet18 Layers} & \textbf{Overflow (\%) using}\\
    & \textbf{$16$-bit accumulator}\\
    \hline   
    \hline
    Convolution layer 2 & 0.0\\
    \hline
    Convolution layer 4 & 0.0\\
    \hline
    Convolution layer 7 & 0.0\\
    \hline
    Convolution layer 9 & 0.0\\
    \hline
    Convolution layer 12 & 0.001\\
    \hline
    %Convolution layer 14 & 0.0027\\
    Convolution layer 14 & 0.003\\
    \hline
    Convolution layer 17 & 0.061\\
    \hline
    Convolution layer 19 & 0.054\\
    \hline
  \end{tabular}
\end{table}
\end{scriptsize}

\begin{scriptsize}
\begin{table}[h!]
  \centering
  \caption{Number of MAC operations performed for generating each output element ($C_{in} \times w \times h$) of different layers of ResNet18 architecture.}
  \label{table:resnet18_ops_count}
  \begin{tabular}{|l|l|l|l|l|l|}
    \hline
    \textbf{ResNet18 Layers} & \textbf{$C_{out}$} & \textbf{$C_{in}$} & \textbf{$w$} & \textbf{$h$} & \textbf{MAC operations}\\
    \hline  
    \hline
    Convolution layer 2 & 64 & 64 & 3 & 3 & 576\\
    \hline
    Convolution layer 4 & 64 & 64 & 3 & 3 & 576\\
    \hline
    Convolution layer 7 & 128 & 128 & 3 & 3 & 1152\\
    \hline
    Convolution layer 9 & 128 & 128 & 3 & 3 & 1152\\
    \hline
    Convolution layer 12 & 256 & 256 & 3 & 3 & 2304\\
    \hline
    Convolution layer 14 & 256 & 256 & 3 & 3 & 2304\\
    \hline
    Convolution layer 17 & 512 & 512 & 3 & 3 & 4608\\
    \hline
    Convolution layer 19 & 512 & 512 & 3 & 3 & 4608\\
    \hline
  \end{tabular}
\end{table}
\end{scriptsize}

%% file: ml_accelerator.tex
\section{Suitability to Hardware Accelerators for Deep Neural Networks}
\label{sec:ml_accelerator}

%applicability to TPU

This section shows an example of how processing circuitry designed for performing the MAC operations in state-of-the-art DNN hardware accelerators can be adapted to support the proposed matrix multiplication instruction between asymmetric bit-width operands. A convolutional operation in DNN layers are typically implemented by lowering 2D convolution to general matrix multiply (GEMM) kernels, which are typically the runtime bottleneck when executed on CPUs, motivating hardware acceleration. Spatial architectures are a class of accelerators that can exploit high compute parallelism of GEMM kernels using direct communication between an array of relatively simple processing engines (PEs). The systolic array (SA) is a coarse-grained spatial architecture for efficiently accelerating GEMM. The SA consists of an array of MAC processing elements (PEs), which communicate operands and results using local register-to-register communication only, which makes the array very efficient and easily scalable without timing degradation. 
These advantages have led to their deployment in commercial products, e.g., the Google Tensor Processing
Unit (TPU)~\cite{tpu2017}.

%A convolutional operation in DNN layers are typically implemented by lowering 2D convolution to general matrix multiply (GEMM) kernels, which are typically the runtime bottleneck when executed on CPUs, motivating hardware acceleration. Spatial architectures are a class of accelerators that can exploit high compute parallelism of GEMM kernels using direct communication between an array of relatively simple processing engines (PEs).
%The systolic array (SA) is a 
%%special-purpose processor
%coarse-grained spatial architecture for efficiently accelerating GEMM. The SA consists of an array of MAC processing elements (PEs), which communicate operands and results using local register-to-register communication only, which makes the array very efficient and easily scalable without timing degradation.

The proposed matrix multiplication instruction at different vector widths (e.g., $128$-bit vector width, etc. as shown in the examples above) will not only play a vital role in offering $2\times$ improvement in throughput of matrix multiplication involving $4$-bit weights and $8$-bit activations in future CPUs, but also will be effective to support MAC operation between $8$- and $4$-bit operands in state-of-the-art DNN hardware accelerators (e.g., TPU, etc.) and offer similar improvement in matrix multiply performance \textit{seamlessly} without violating the various implementation constraints.

%The proposed matrix multiplication instruction 
%%between asymmetric bit-width operands 
%at different vector widths (e.g., $128$-bit vector width, etc. as shown in this work) will not only play a vital role in offering $2\times$ improvement in throughput of matrix multiplication involving $4$-bit weights and $8$-bit activations in future CPUs, but also will be quite %%useful
%effective to support MAC operation between $8$- and $4$-bit operands in state-of-the-art DNN hardware accelerators (e.g., TPU, etc.) and offer similar improvement in matrix multiply performance \textit{seamlessly} without violating the various implementation constrains.
%(identical number of rows and columns in accumulator register tile of Mortlach architecture). 

\begin{figure}[!tbp]
  \centering
  \includegraphics[width=1.0\linewidth]{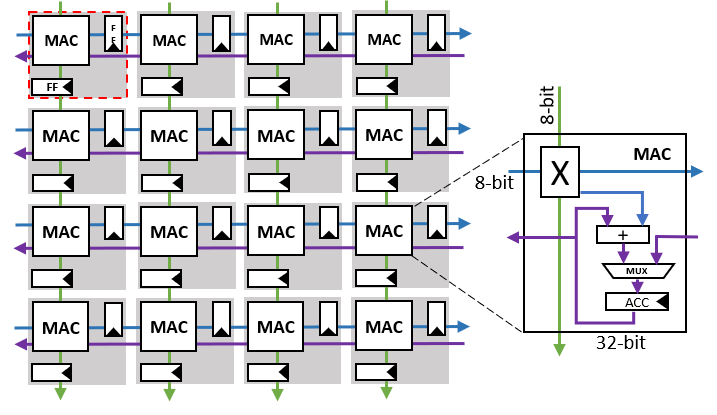}
  \caption{ Conventional systolic array (SA) microarchitectures. Pipeline registers connect adjacent PEs, with only local data movement.}
  \label{fig:SA_int8_operands}
\end{figure}

Figure~\ref{fig:SA_int8_operands} shows the structure of a SA widely deployed in Google TPUs. It is designed for supporting multiplications involving operands with equal element size.  Each MAC operation in the SA requires two $8$-bit operand registers. The $16$-bit products are collected into the $32$-bit accumulator buffers. This SA organization enables output-stationary dataflow, which keeps the larger $32$-bit accumulators in place and instead shifts the smaller $8$-bit operands.
%Figure~\ref{fig:SA_int8_operands} shows the structure of a conventional SA, widelydeployed  in  Google  TPUs. Each MAC operation in the SA of the TPUs requires two INT8 operand registers. The $16$-bit products are collected into the $32$-bit accumulator buffers. This SA organization enables output-stationary dataflow, which keeps the larger INT32 accumulators in place and instead shifts the smaller INT8 operands.

%Each MAC operation traditionally requires two INT8 operand registers and one INT32 accumulator.
%The MACs in the Google TPU performs $8$-bit multiply-and-adds on signed or unsigned integers, and the $16$-bit products are collected into the $32$-bit accumulators below the matrix unit (register file)

\begin{figure}[!tbp]
  \centering
  \includegraphics[width=1.0\linewidth]{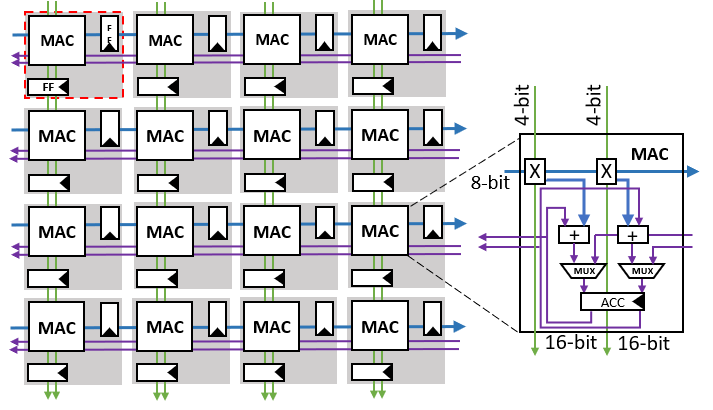}
  \caption{Systolic array (SA) microarchitectures, modified by extending each scalar PE to perform two MAC operations between $8$-bit activations and $4$-bit weights, while accumulating the product into $16$-bit accumulator buffers.}
  \label{fig:SA_int8int4_operands}
\end{figure}

Figure~\ref{fig:SA_int8int4_operands} shows how a MAC operation acting on $8$-bit and $4$-bit operands can be performed using a SA architecture. The $8$-bit operand registers now can accommodate two $4$-bit weight values and a MAC unit now can perform two MACs between $8$-bit and $4$-bit operands values to generate two $12$-bit products. The $12$-bit products in turn are accumulated into $16$-bit accumulators, thus enabling the $32$-bit accumulator buffer of the SA of Figure~\ref{fig:SA_int8_operands} to be re-purposed for collecting two $16$-bit wide MAC output values. Thus the MAC operation between $8$-bit and $4$-bit operands generating $16$-bit output values can be seamlessly integrated into the SA matrix multiplication engine to achieve $2\times$ improvement in MAC throughput without violating the implementation constraints around the size of operand buffers and accumulator buffers. Similarly, a SA architecture that enforces weight-stationary dataflow can easily be extended to support the proposed matrix multiplication operation involving asymmetric bit-width operands. Weight-stationary dataflow keeps the smaller $8$-bit weights in place and shifts the larger $32$-bit accumulator values.
%Figure~\ref{fig:SA_int8int4_operands} shows the mapping of a MAC operation between $8$- and $4$-bit operands into the conventional SA architecture. A $8$-bit operand registers now can accommodate two $4$-bit weight values and a MAC unit now can perform two multiply-and-adds between $8$- and $4$-bit operands values to generate two $12$-bit products.
%(and storing the output value into a $16$-bit accumulator operand buffer) 
%shows the dataflow of a 4x4 mac operation between 8 and 4 bit operands
%into the TPU SA matrix multiplication engine/array structure.
%The $12$-bit products in turn are accumulated into $16$-bit accumulators, thus enabling the $32$-bit accumulator buffer of a MAC unit of conventional SA to be re-purposed for collecting two $16$-bit wide MAC output values.

%Thus the MAC operation between $8$- and $4$-bit operands generating $16$-bit output values can be seamlessly integrated into the conventional SA matrix multiplication engine to achieve $2\times$ improvement in MAC throughput without violating the implementation constraints around the size of operand buffers and accumulator buffers.

%SA accumulator register file (identical number of rows and columns in the accumulator register tile of Mortlach architecture).

%Similarly, a SA architecture that enforces weight-stationary dataflow can easily be extended to support the proposed matrix multiplication operation involving asymmetric bit-width operands. Weight-stationary dataflow keeps the smaller INT8 weights in place and shifts the larger INT32 accumulator values.

%applicability to eyeriss

%% file: conclusion.tex
\section{Conclusion and Future Work}
\label{sec:conclusion}

We propose a SIMD matrix multiplication operation to obtain $2\times$ increase in MAC throughput for asymmetric bit-width operands without requiring either additional register read and write ports in CPUs or larger operand and accumulator buffers in DNN accelerators.
The matrix multiply instruction makes this possible by accumulating product values into $16$-bit accumulators 
%at 128-bit vector width
as opposed to $32$-bit accumulators used for symmetric $8$-bit operands. 
%to meet the register vector width requirements in CPUs or implementation constraints of DNN accelerators.
We observed negligible overflow ($0.05\%$) from $16$-
bit accumulators for the pre-trained ResNet-18 model with $4$-bit weights and $8$-bit activations.
A natural next step is to explore the impact of this negligible overflow on the accuracy of the pre-trained ResNet-18 model. We believe this $0.05\%$ overflow from narrower $16$-bit accumulators can be avoided via integrating the constraint on accumulator's width into the training procedure of the ResNet-18 model with $4$-bit weights. We leave this exploration for future work.
In future, we plan to explore the impact of $4$-bit weights and $16$-bit accumulators on other highly optimized CNNs, especially MobileNets. In addition, it will be interesting to see how the theoretical gains reported here from asymmetric bit-width operands translate into actual energy savings and runtime speedups on DNN accelerator and CPU simulators~\cite{gem5_simulator}.

%This is a proposal for a new matmul instruction that uses mixed precision on its inputs (1 8-bit and 1 4-bit argument) and accumulates into a narrower 16-bit output for future heavily quantized machine learning models. 

%ISA extension for supporting future ML workloads at high compute throughput/balanced utilization of register port bandwidth 

%Advantage(s) over what is currently being done: 
%Provides advantages of high throughput benefit without requiring additional register read ports. 